\documentclass[lettersize,journal]{IEEEtran}
\usepackage{amsmath,amsfonts}
\usepackage{algorithmic}
\usepackage{algorithm}
\usepackage{array}
\usepackage{textcomp}
\usepackage{stfloats}
\usepackage{url}
\usepackage{verbatim}
\usepackage{graphicx}
\usepackage{cite}
\usepackage{tabularx}
\usepackage{booktabs}
\usepackage{multirow}
\usepackage{subfigure}

\begin{document}

\title{Detecting AutoEncoder is Enough to Catch LDM Generated Images}

\author{Dmitry Vesnin,
        Dmitry Levshun,
        Andrey Chechulin
\thanks{All authors are from St. Petersburg Federal Research Center of the Russian Academy of Sciences, Saint-Petersburg, Russia.}
\thanks{Corresponding author: Dmitry Levshun (levshun.d@iias.spb.su).}

}


\maketitle

\begin{abstract}
In recent years, diffusion models have become one of the main methods for generating images. However, detecting images generated by these models remains a challenging task. This paper proposes a novel method for detecting images generated by Latent Diffusion Models (LDM) by identifying artifacts introduced by their autoencoders. By training a detector to distinguish between real images and those reconstructed by the LDM autoencoder, the method enables detection of generated images without directly training on them. The novelty of this research lies in the fact that, unlike similar approaches, this method does not require training on synthesized data, significantly reducing computational costs and enhancing generalization ability. Experimental results show high detection accuracy with minimal false positives, making this approach a promising tool for combating fake images.
\end{abstract}


\section{Introduction}
\label{sec:intro}
In the past few years, there has been significant progress in image generation. New diffusion models such as DDPM \cite{ho2020denoising} and DDIM \cite{songdenoising} have outperformed the previous state-of-the-art approach, GANs, in image generation \cite{dhariwal2021diffusion} and introduced a new paradigm for generative modeling.

Although these diffusion-based models achieve better quality for generated images, they have a significant drawback – they require extensive computational resources for training and generation. Additionally, they cannot be easily scaled to generate high-resolution images because they operate in a space with numerous dimensions – the pixel space.

In \cite{rombach2022high} the authors proposed a new method called LDM (Latent Diffusion Model), which operates in a much smaller latent space instead of the pixel space. By conditioning the generation process on text, it became possible to generate images based on text prompts. LDM trained and generated new samples much faster, and it became a standard for text-to-image generation.

Using large datasets of images and text, such as LAION-5B \cite{schuhmann2022laion}, companies began to introduce image generation capabilities into their products. Examples include MidJourney, DALLE 3, and Adobe FireFly, which allow non-technical users to create images using text queries. Some companies provide their models and code for training as open-source, such as Stable Diffusion. This enables enthusiasts to customize these models and disable built-in watermarks. The rise of user-friendly graphical interfaces for image generation, such as ComfyUI\cite{comfyanonymous2024comfyui} and stable-diffusion-webui\cite{automatic11112022stable}, further lowers the entry barrier for creating images. These tools allow users to generate images autonomously without relying on third parties.

\begin{figure}[h]
    \centering
    \includegraphics[width=0.9\columnwidth]{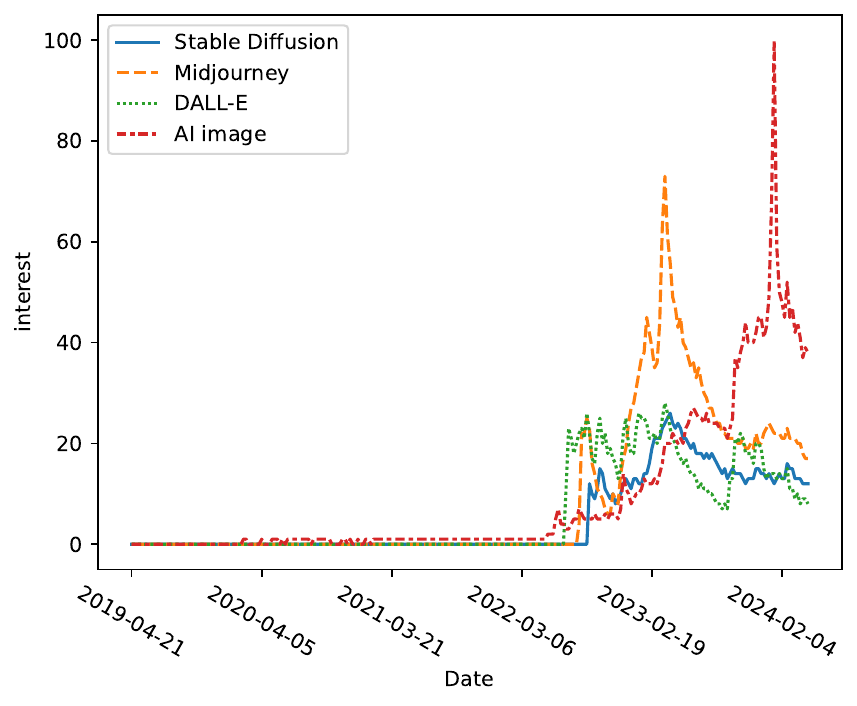}
    \caption{Popularity graph of image generation-related search queries on Google}
    \label{fig:trends}
\end{figure}

In 2022, OpenAI reported that users were creating more than 2 million images daily using DALL-E \cite{openai_dalle_2022}, and in 2023, Adobe reported that Adobe FireFly generated over 1 billion images in 3 months after its launch \cite{adobe_firefly_2023}. Figure~\ref{fig:trends} shows the trends of search queries related to image generation.

To prevent the spread of misinformation through generated images and to properly label them, reliable detectors are needed, with a focus on minimizing false positives.

This study presents a simple yet effective approach for detecting images generated by latent diffusion models. We hypothesize that artifacts introduced by the encoder and decoder of the autoencoder, which projects images from the RGB pixel space to a lower-dimensional latent space and back, are sufficient for detecting images generated by latent diffusion models. This also simplifies the detection of manipulations such as inpainting (partial image editing), as they require projecting images into the latent space.

Unlike other methods, our solution leverages artifacts introduced by the autoencoder encoder and decoder to effectively detect LDM-generated images without the need for training on the generated images themselves. The proposed method shows high generalization, resilience to distortions such as JPEG compression, and significantly reduces computational costs, making it an efficient tool for detecting images created by various latent diffusion models. This constitutes the scientific novelty and practical significance of the proposed solution.

The paper is organized as follows: Section \ref{sec:related} provides an overview of the current state of research and practical results in detecting diffusion model-generated images. Section \ref{sec:solution} describes the proposed solution, a method for detecting latent diffusion models without training on generated images. Section \ref{sec:evaluation} presents experimental results demonstrating the effectiveness of the proposed approach. Section \ref{sec:disc} offers a brief analysis of the strengths and weaknesses of the solution. Finally, the conclusion summarizes the findings and suggests future research directions.

\section{Related Work}
\label{sec:related}
In recent years, there has been a significant increase in interest in the creation and detection of synthetic images generated by diffusion models such as DDPM and its modifications. Along with this, methods for detecting such images have also been evolving, including deep learning-based approaches that aim to identify artifacts arising during the generation process.

Diffusion models presented in \cite{ho2020denoising} generate new synthetic images by learning to transform a simple distribution (usually Gaussian) into a target distribution, such as the distribution of images, using a Markov chain of diffusion steps. The forward diffusion process takes a sample from the target distribution and gradually adds Gaussian noise to it over T steps. The reverse diffusion process tries to predict the added noise from the noisy sample. By learning to remove the noise added to the image, the model learns to transform the Gaussian distribution into the target distribution.

The neural network architecture typically used for diffusion models, such as DDPM, is U-Net \cite{ronneberger2015u}. However, replacing it with a transformer architecture is an active area of research, for example, DiT \cite{peebles2023scalable}.

\begin{figure}[h]
    \centering
    \includegraphics[width=\columnwidth]{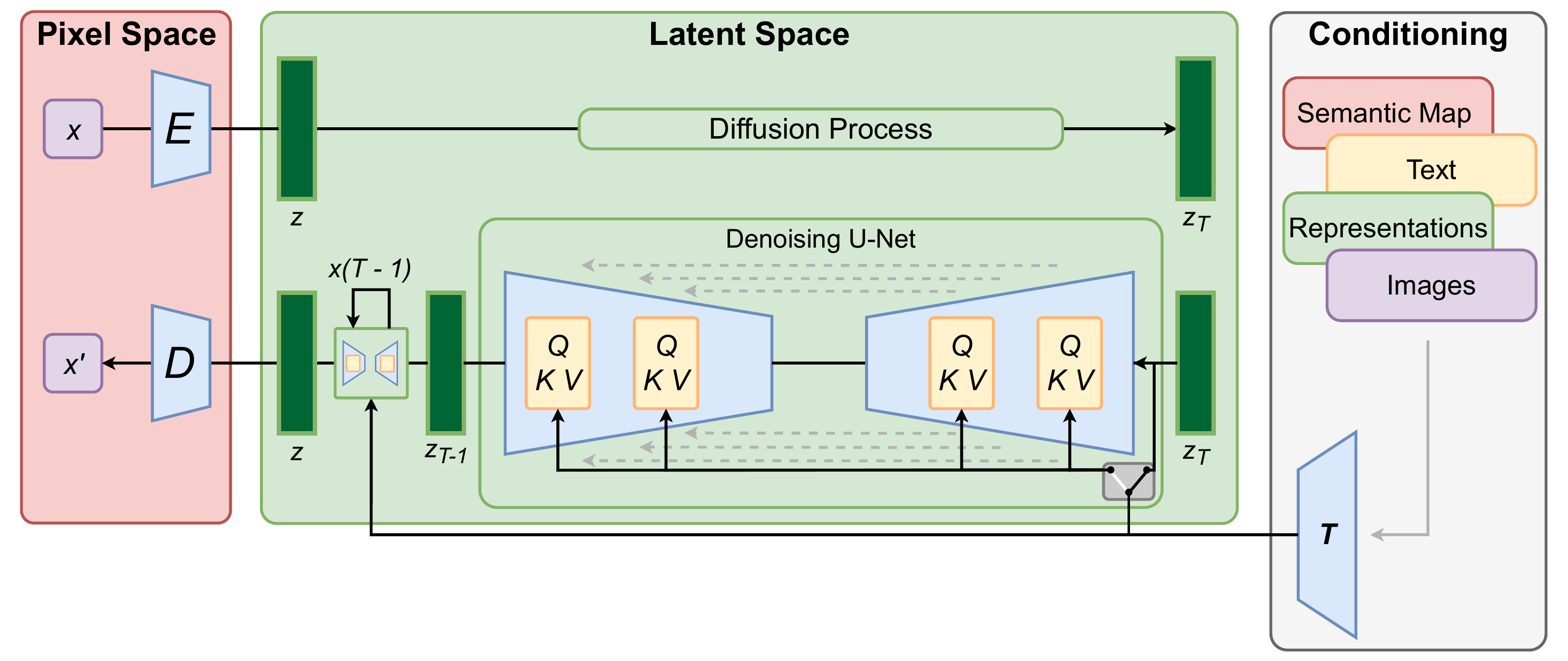}
    \caption{LDM Architecture Schematic}
    \label{fig:arch}
\end{figure}

Although DDPM and its modifications outperform GANs, they have a major drawback: slow training and generation. Working in the high-dimensional RGB pixel space is very computationally expensive. In LDM, the authors use a pre-trained autoencoder to transform images into a low-dimensional latent space and perform diffusion in that space. A schematic architecture of LDM is shown in Figure \ref{fig:arch}.

The model is trained in two stages: in the first stage, the autoencoder is trained to compress images. Since the latent space is much smaller than the pixel space, the autoencoder ignores high-frequency details and learns to efficiently compress images. In the second stage, a diffusion model is trained to generate new samples in the latent space, which are then decoded by the pre-trained decoder \cite{rombach2022high}.

However, as mentioned above, alongside the development of diffusion models, interest in methods for detecting such images is also increasing, which is particularly important in combating fake images. In recent years, several key scientific trends have emerged in this area. 

First, deep learning methods, such as convolutional neural networks (CNNs) and vision transformers, are actively being developed to detect specific artifacts that arise when creating fakes, including deepfakes \cite{ha2023robust,aghasanli2023interpretable}. These methods are capable of identifying small inconsistencies, such as skin texture artifacts, compression artifacts, anomalies in reflections or shadows, and changes in facial structure. 

Second, models are being developed not only to detect fakes but also to explain which parts of the image raised suspicion, which is important for interpreting the results. 

Third, significant attention is being given to creating universal models that can effectively handle new types of fakes with minimal retraining, which is especially relevant given the rapid emergence of new methods for creating fakes. 

Finally, research is being conducted on protecting models from attacks aimed at bypassing them and improving the resilience of detection algorithms to various forms of adversarial manipulation by attackers.

One key research direction involves adapting existing methods developed for detecting images generated by Generative Adversarial Networks (GANs) to new types of generation based on diffusion models. Additionally, universal detectors are being developed that can effectively recognize fake images regardless of the generative model used, by focusing on features specific to AI-generated images. Particular attention is being paid to methods that do not require training on synthetic images, which significantly improves the generalization capability of detectors and reduces computational costs. Furthermore, efforts are being made to develop methods that maintain their effectiveness under various image distortions, such as JPEG compression or resolution changes, which is crucial for real-world applications.

In \cite{corvi2023detection}, the authors evaluate whether models trained to detect GAN-generated images can detect images created using LDMs. The performance of models trained on GAN-generated images decreases; however, adding diffusion-generated images to the training dataset improves it.

In \cite{ojha2023towards}, the development of a universal fake image detector is considered, which can determine whether an arbitrary image is fake, with access to only one type of generative model during training. The authors first demonstrate that the existing paradigm of training a deep network to classify real and fake images does not allow for detecting fake images generated by diffusion models when trained to detect fake images generated by GANs. They found that the resulting classifier is asymmetrically tuned to detect patterns that make an image fake. Based on this discovery, they propose performing classification of real and fake images without training, using a feature space that has not been explicitly trained to distinguish real images from fake ones.

This idea is implemented in two ways: through K-Nearest Neighbor classification and with a single trainable linear layer with a sigmoid activation function for binary classification. To extract image features, the authors use the CLIP \cite{radford2021learning} ViT-L/14 model. The training dataset is generated by the ProGAN model. With access to the feature space of a large pre-trained model like CLIP, even a simple k-NN classifier has strong generalization capabilities and outperforms models trained from scratch.

In \cite{ricker2022towards}, the authors show that detectors trained on GAN-generated images perform poorly on images generated by diffusion models. However, detectors trained on diffusion-generated images perform much better at detecting GAN-generated images. Diffusion models have fewer artifacts in the frequency domain but underestimate higher frequencies in the spectrum, which may be caused by the loss function used for training diffusion models.

In \cite{wang2023dire}, it was found that images generated by diffusion models can be reconstructed more accurately than the original images. By training a binary classifier based on DIRE features, which represent the difference between an image and its reconstructed version, better accuracy is achieved than when using RGB images.

In \cite{ricker2024aeroblade}, a detection method without training is presented. Unlike DIRE, this method uses the autoencoder from a latent diffusion model to reconstruct the image. By measuring the LPIPS metric \cite{zhang2018unreasonable} between the original image and its version reconstructed by the LDM autoencoder, it is possible to determine whether the image was generated by LDM models, as they have a smaller reconstruction error. By using different autoencoders, different models can be detected.

The summary of the recent methods analysis and their comparison with the proposed one is presented in Table \ref{tab:comparison-methods}.

\begin{table}[h]
\centering
\caption{Comparison of methods}
\label{tab:comparison-methods}
\begin{tabular}{@{}ccccc@{}}
\toprule
\textbf{Method} & \textbf{Training} & \textbf{Synthetic Data} & \textbf{LDM autoencoder} \\ \midrule
\cite{corvi2023detection} & Required & Used & Not used \\
\cite{ojha2023towards} & Required & Used & Not used \\
\cite{wang2023dire} & Required & Used & Not used \\
\cite{ricker2024aeroblade} & Not required & Not used & Inference \\
\textbf{Proposed} & Required & Not used & Training \\ \bottomrule
\end{tabular}%
\end{table}

It shows that despite the availability of solutions for detecting fake images, current methods do not provide sufficient accuracy when detecting images generated by diffusion models, especially in the presence of distortions or changes in the model. Moreover, existing solutions depend on the quality of the training dataset used for training the detector model. Thus, there is a need for a new approach that can effectively detect generated images without the necessity of training on synthetic data. This paper presents research aimed at solving this problem by developing a method that leverages artifacts introduced by latent diffusion model autoencoders, which improves the accuracy of detecting generated images.

\section{Suggested Solution}
\label{sec:solution}
This section describes in more detail the approach proposed in this paper for detecting generated images. Specifically, Section \ref{subsec:problem} outlines the problem statement, Section \ref{subsec:approach} discusses the main stages of the approach, and Section \ref{subsec:metrics} describes the key metrics used for evaluation.

\subsection{Problem Statement}
\label{subsec:problem}
This paper addresses the problem of detecting reconstructed or generated images. The input for this task is a set of image crops $IC$ of the analyzed image:

\[
IC = \{ic_1, ic_2, ..., ic_n\}, \quad n \in \mathbb{N}
\]
where $ic_j$ is the $j$-th crop of the analyzed image, $j \in \{1..n\}, n \in \mathbb{N}$.

Thus, the task of detecting reconstructed or generated images can be formulated as follows:

\[
f(IC, m, t) = 
\begin{cases} 
1, & \text{if the image is not original}; \\
0, & \text{if the image is original}, 
\end{cases}
\]
where $f$ is the detection function for image reconstruction or generation, taking as input $IC$, $m$, and $t$, and providing a binary decision as output; $m$ is the AI model used; $t$ is the probability threshold for decision-making, which is set individually for each model depending on the number of crops.

\subsection{Approach}
\label{subsec:approach}
\begin{figure}[h]
    \centering
    \includegraphics[width=\columnwidth]{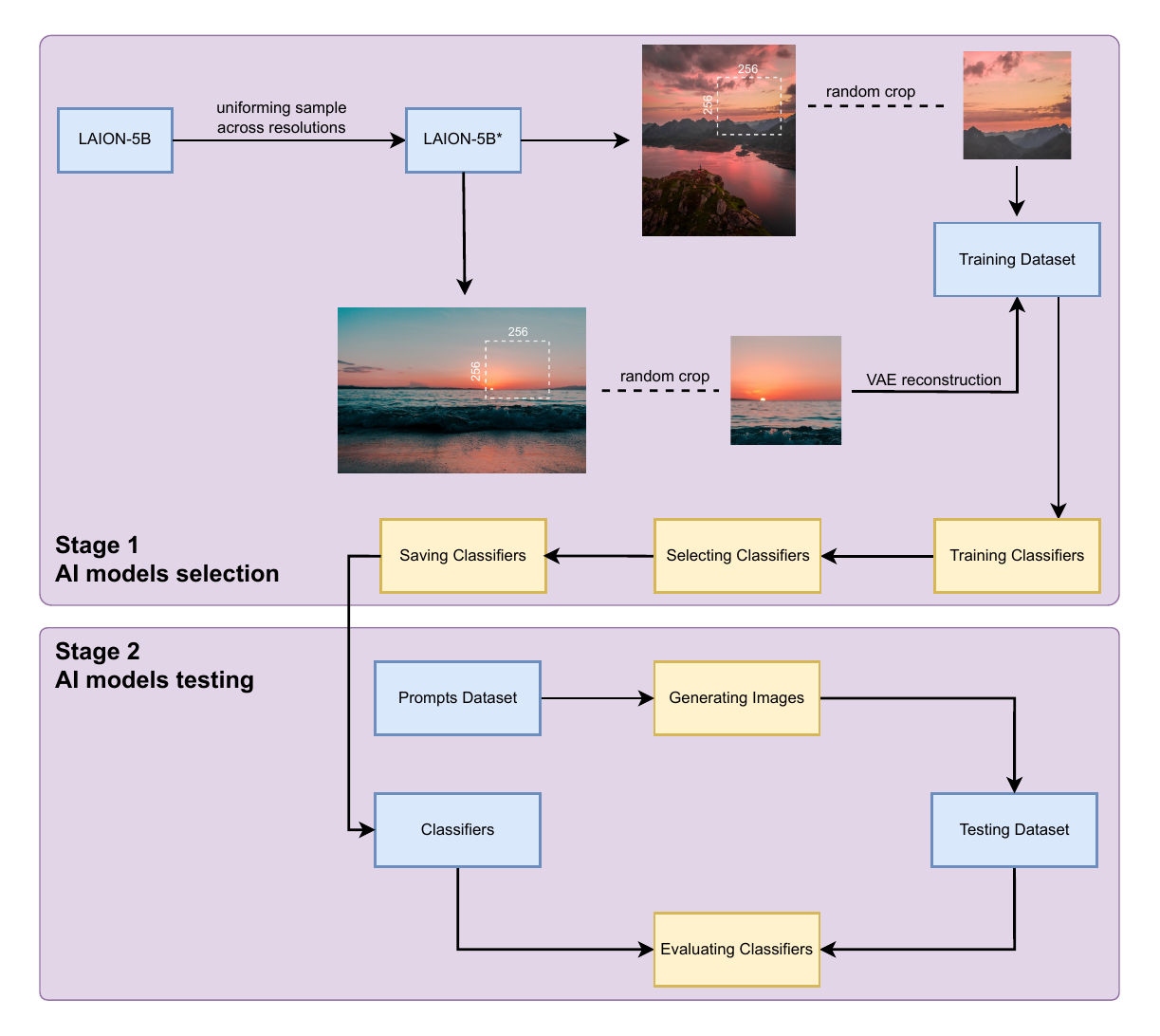}
    \caption{The approach for detecting generated images}
    \label{fig:approach-detection}
\end{figure}

The approach for detecting generated images is illustrated in Figure~\ref{fig:approach-detection} and consists of two stages: selection of AI models and their testing.

\textit{Stage 1: AI models selection.} This stage has two key steps. Let us consider them in detail.

\textit{Step 1.1: Dataset Preparation.} At this step, two tasks are solved – the preparation of a dataset for model training and the preparation of a dataset for its evaluation.

The core dataset for detecting reconstructed images is the LAION-5B, which contains 5 billion pairs of images and texts. In this work, we use only the images and discard the captions.

Most images available on the internet are of low or medium resolution. To address this issue, we adjusted the distribution by intentionally selecting images with varying resolutions. For this purpose, the images were divided into 14 groups based on the number of pixels from 300² to 6000²: (224-300, 300-400, 400-500, 600-700, 800-900, 900-1000, 1000-1500, 1500-2000, 2000-2500, 3000-3500, 4000-6000).

After downloading the images, random crops of 256x256 pixels were extracted. Then, reconstructed versions of the images were obtained by encoding and decoding them using the VAE from Stable Diffusion 2.1.

For testing, a separate dataset of generated images was collected using the following models:
\begin{itemize}
    \item \texttt{cagliostrolab\_animagine\_xl\_3.1}
    \item \texttt{CompVis\_stable\_diffusion\_v1\_4}
    \item \texttt{CompVis\_ldm\_text2im\_large\_256}
    \item \texttt{dreamlike\_art\_dreamlike\_photoreal\_2.0}
    \item \texttt{stabilityai\_stable\_diffusion\_2}
    \item \texttt{kandinsky\_community\_kandinsky\_3}
    \item \texttt{stabilityai\_stable\_diffusion\_2\_1}
    \item \texttt{runwayml\_stable\_diffusion\_v1\_5}
    \item \texttt{ckpt\_anythingV3}
    \item \texttt{stabilityai\_stable\_diffusion\_xl\_base\_1.0}
    \item \texttt{facebook\_DiT\_XL\_2\_512}
    \item \texttt{MidJourney v6}
\end{itemize}

Images were generated using prompts from \cite{Gustavosta_Stable_Diffusion_Prompts}.
The models cover a wide range of different architectures:
\begin{itemize}
    \item LDM from the original Stable Diffusion paper, Stable Diffusion XL \cite{podell2023sdxl};
    \item  DiT (a latent diffusion model that uses transformers instead of U-Net);
    \item Kandinsky 3 (which uses MoVQ instead of VAE);
    \item user-trained models such as AnythingV3, Animagine, and Dreamlike Photoreal.
\end{itemize}

Note that not all the listed models use the VAE that we used for image reconstruction in our training dataset. Working with such a testing data allows one to assess the robustness and generalizability of the proposed solution.

Stable Diffusion 1.4, 1.5, and CompVis ldm\_text2im\_large\_256 share the same encoder and decoder. Stable Diffusion 2 and 2.1 also use the same encoder and decoder, with the encoder identical to that used in SD 1.4. DiT\_XL\_2\_512 utilizes the autoencoder from Stable Diffusion 2. The remaining models differ from each other in both their encoders and decoders.

\textit{Step 1.2: AI models evaluation. }At this step, two architectures were used for the detectors: CNN and ViT, specifically ConvNext (CNN) \cite{liu2022convnet} and EVA-02 (ViT) \cite{fang2024eva}.

To improve overall accuracy, models pre-trained on the CLIP task (Contrastive Language-Image Pre-Training) using large datasets like LAION-2B and Merged-2B were selected.

Additionally, a small CNN model, EfficientNet-V2 B0 \cite{tan2021efficientnetv2}, pre-trained on ImageNet-1k, was chosen to assess the differences in detection performance between simpler and more complex models.

We trained the described models on the binary classification task using the binary cross-entropy loss function. We classified the crops of original images and crops of images reconstructed by the autoencoder from SD 2.1. Afterward, we evaluated how well the models learned to detect such reconstructed images. Then, we tested the models on a dataset consisting of both generated and original images to assess the models' ability to detect generated images.

\textit{Stage 2: AI models testing.} The process of detecting generated images also consists of two key steps aimed at data preprocessing and decision-making. Let’s examine each step of this process in more detail.

\textit{Step 2.1: Data preprocessing.} At the beginning of this step, a crop or several crops of the required size are randomly selected from the image. The obtained crops are then passed to the models trained in the previous stage to determine whether each individual crop is generated.

\textit{Step 2.2: Analyzing crops and making decisions.} At this stage, each model outputs the probability of the image to be generated. The output ranges from 0 to 1, with crops considered generated, when probability exceeds 0.5.

The first method developed for inference decides whether an image is generated based on the analysis of just one crop. This method is referred to as ``1 try''.

An alternative method for inference has also been developed, where the decision is made based on the analysis of 10 random crops. In this case, the final probability compared to the threshold value is the average of the probabilities obtained for all analyzed crops. This method is called ``10 tries''.

For the ``10 tries'' method, threshold values were established for each model to minimize false positives without significantly affecting recall:
\begin{itemize}
    \item ConvNext Large - 0.20;
    \item EVA-02 ViT L/14 - 0.30;
    \item EfficientNet-V2 B0 - 0.50.
\end{itemize}

We propose a method that enables the detection of LDMs without training on generated images. The task is formulated as a binary classification problem with two classes: original images and images reconstructed using the autoencoders from LDM. As will be demonstrated later, this is sufficient for detecting images generated by LDMs.

\subsection{Evaluation Metrics}
\label{subsec:metrics}
To evaluate the detectors, the following classification metrics were used: precision, recall, and F1-score. These metrics are defined as follows:

\[
\text{Precision} = \frac{TP}{TP + FP}, \text{Recall} = \frac{TP}{TP + FN},
\]
\[
\text{F1} = 2 \times \frac{\text{Precision} \times \text{Recall}}{\text{Precision} + \text{Recall}}
\]
where $TP$ is the number of true positives (correctly detected reconstructed images), $FP$ is the number of false positives (original images incorrectly detected as reconstructed), and $FN$ is the number of false negatives (reconstructed images incorrectly detected as original).

For evaluating the detectors on the task of detecting generated images, we used the True Positive Rate (TPR) and False Positive Rate (FPR). TPR is also known as recall or sensitivity. FPR measures the rate at which original images are incorrectly classified as generated:

\[
\text{TPR} = \frac{TP}{TP + FN}, \text{FPR} = \frac{FP}{FP + TN}
\]
where $TN$ represents true negatives (correctly identified original images).

\section{Experimental Evaluation}
\label{sec:evaluation}
This section presents the experimental results of our method. Section \ref{subsec:datasets} describes the datasets used, Section \ref{subsec:setup} outlines the experimental setup, and Section \ref{subsec:results} presents the results.

\subsection{Datasets}
\label{subsec:datasets}
By applying the methods described in Section \ref{sec:solution} to the LAION-5B dataset, we obtained 3 million images:
\begin{itemize}
    \item 1 million crops of original images with normal resolution (gathered using sampling described in Section \ref{sec:solution});
    \item 1 million reconstructed crops;
    \item 1 million crops of original images with high resolution (larger than 2500x2500 pixels).
\end{itemize}

The dataset was split in a 75/25 ratio into training and test. These splits were used for training models and assessing their ability to detect reconstructed images.

To test the models’ ability to detect generated images, we applied the methods described in Section \ref{sec:solution} and collected a dataset of generated images, see Table~\ref{table:gen-data}.

\begin{table}[h]
\centering
\caption{Dataset of Generated Images}
\label{table:gen-data}
\begin{tabular}{@{}lc@{}}
\toprule
\multicolumn{1}{c}{\textbf{Model}} & \textbf{Number of Images} \\ \midrule
cagliostrolab\_animagine\_xl\_3.1 & 1,000 \\
CompVis\_stable\_diffusion\_v1\_4 & 1,000 \\
CompVis\_ldm\_text2im\_large\_256 & 1,000 \\
dreamlike\_art\_dreamlike\_photoreal\_2.0 & 1,000 \\
stabilityai\_stable\_diffusion\_2 & 1,000 \\
kandinsky\_community\_kandinsky\_3 & 1,000 \\
stabilityai\_stable\_diffusion\_2\_1 & 1,000 \\
runwayml\_stable\_diffusion\_v1\_5 & 1,000 \\
anythingV3 & 1,000 \\
stabilityai\_stable\_diffusion\_xl\_base\_1.0 & 1,000 \\
facebook\_DiT\_XL\_2\_512 & 1,000 \\
MidJourney\_v6 & 1,000 \\ \midrule
\textbf{Total} & 12,000 \\ \bottomrule
\end{tabular}
\end{table}

Additionally, we generated two datasets with original images using LAION-5B. They contain:
\begin{itemize}
    \item \texttt{original} – a set of original images (gathered using sampling described in Section \ref{sec:solution}), 36~000 images;
    \item \texttt{high\_res\_imgs} – images with a resolution greater than 2500x2500 pixels, 15~000 images.
\end{itemize}

These datasets were necessary for evaluating the false positive rate of our models.

\subsection{Setup}
\label{subsec:setup}
Experiments were conducted using NVIDIA GeForce RTX 3090 GPU videocard with 24GB of memory, Intel Core i5-13600KF processor running at 5.10 GHz with 14 cores, and 64GB of DDR4 RAM.

For training, we used the AdamW optimizer with 10 epochs and a weight decay equal to 0.05. A learning rate scheduler was applied with linear warm-up for 5~000 steps, followed by cosine decay, starting from an initial learning rate of $5 \times 10^{-6}$.

To verify that the models could detect reconstructed images:
\begin{itemize}
    \item EfficientNet-V2 B0 (5.8 millions of parameters) and ConvNext Large (198.5 millions of parameters) were trained on images of size 256x256;
    \item EVA-02 ViT L/14 (303.3 millions of parameters) was trained on images of size 252x252, as image sizes must be divisible by the patch size (14).
\end{itemize}

After training the models on the task of detecting reconstructed images, we tested them on the dataset of generated images to evaluate their ability to detect generated images.

\subsection{Results}
\label{subsec:results}
Table~\ref{tab:models-results} shows the results on the test set consisting of original and reconstructed images. It can be observed that the ConvNext Large model achieved the best results, but all models demonstrated relatively similar performance.

\begin{table}[h]
\caption{Testing results on original and reconstructed images}
\label{tab:models-results}
\begin{tabular}{@{}llccc@{}}
\toprule
\multicolumn{1}{c}{\textbf{Model}} & \multicolumn{1}{c}{\textbf{Class}} & \textbf{Precision} & \textbf{Recall} & \textbf{F1-Score} \\ \midrule
\multirow{2}{*}{EfficientNet-V2 B0} & Original & 0.96 & 0.98 & 0.97 \\
 & Reconstructed & 0.95 & 0.92 & 0.93 \\ \midrule
\multirow{2}{*}{EVA-02 ViT L/14} & Original & 0.98 & 0.99 & 0.99 \\
 & Reconstructed & 0.98 & 0.95 & 0.97 \\ \midrule
\multirow{2}{*}{ConvNext Large} & Original & 0.98 & 1.00 & 0.99 \\
 & Reconstructed & 1.00 & 0.96 & 0.98 \\ \bottomrule
\end{tabular}
\end{table}

The experimental results show that all models successfully learned to detect reconstructed images. Next, we evaluate whether this ability is sufficient to detect generated images, see Table~\ref{tab:gen-results}. The table shows the TPR for each model across different tested image categories. We evaluated the detectors' performance with two methods: ``1~try'' (decision based on a single crop) and ``10~tries'' (decision based on the average result of 10 random crops).

\begin{table}[h]
\centering
\caption{Experimental Results on Generated Images}
\label{tab:gen-results}
\resizebox{\columnwidth}{!}{%
\begin{tabular}{@{}lccccccc@{}}
\toprule
\multicolumn{1}{c}{\multirow{2}{*}{\textbf{Data Source}}} & \multirow{2}{*}{\textbf{Metric}} & \multicolumn{2}{c}{\textbf{EfficientNet-V2 B0}} & \multicolumn{2}{c}{\textbf{ConvNext Large}} & \multicolumn{2}{c}{\textbf{EVA-02 ViT L/14}} \\ \cmidrule(l){3-8} 
\multicolumn{1}{c}{} &  & \textbf{1 try} & \textbf{10 tries} & \textbf{1 try} & \textbf{10 tries} & \textbf{1 try} & \textbf{10 tries} \\ \midrule
original & FPR & 0.01228 & 0.00147 & 0.00035 & 0.00024 & 0.00359 & 0.00098 \\
high\_res\_imgs & FPR & 0.00196 & 0.00000 & 0.00014 & 0.00000 & 0.00108 & 0.00000 \\
cagliostrolab\_animagine\_xl\_3.1 & TPR & 0.57000 & 0.61000 & 0.86400 & 0.99100 & 0.91100 & 0.99500 \\
CompVis\_stable\_diffusion\_v1\_4 & TPR & 0.62400 & 0.68370 & 0.90990 & 0.98320 & 0.95070 & 0.98840 \\
CompVis\_ldm\_text2im\_large\_256 & TPR & 0.88500 & 0.88500 & 0.99600 & 0.99800 & 0.96800 & 0.99000 \\
dreamlike\_art\_dreamlike\_photoreal\_2.0 & TPR & 0.71100 & 0.78500 & 0.96000 & 0.99500 & 0.98900 & 0.99900 \\
stabilityai\_stable\_diffusion\_2 & TPR & 0.50800 & 0.54600 & 0.98700 & 1.00000 & 0.99300 & 1.00000 \\
kandinsky\_community\_kandinsky\_3 & TPR & 0.59500 & 0.67800 & 0.89500 & 0.99400 & 0.81400 & 0.98700 \\
stabilityai\_stable\_diffusion\_2\_1 & TPR & 0.54400 & 0.55200 & 0.98000 & 0.99900 & 0.98900 & 1.00000 \\
runwayml\_stable\_diffusion\_v1\_5 & TPR & 0.62100 & 0.65650 & 0.92900 & 0.97800 & 0.95920 & 0.99370 \\
ckpt\_anything\_v3.0 & TPR & 0.86300 & 0.91600 & 0.97200 & 0.99200 & 0.99700 & 1.00000 \\
stabilityai\_stable\_diffusion\_xl\_base\_1.0 & TPR & 0.51800 & 0.56100 & 0.95700 & 1.00000 & 0.93700 & 0.99800 \\
facebook\_DiT\_XL\_2\_512 & TPR & 0.49100 & 0.48000 & 0.91700 & 0.97100 & 0.84300 & 0.91900 \\
MidJourney\_v6 & TPR & 0.57010 & 0.61960 & 0.90420 & 0.98570 & 0.91180 & 0.98650 \\ \bottomrule
\end{tabular}%
}
\end{table}

The experiments showed that EfficientNet-V2 B0 significantly underperforms compared to the other models in detecting generated images. The ConvNext Large and EVA-02 ViT L/14 models demonstrated comparably high detection accuracy. The high level of detection for models not present in the training dataset indicates that they share common artifacts independent of the model architecture. All tested models used autoencoders for projecting images into latent space and back. Kandinsky 3 uses MoVQ \cite{zheng2022movq}, while the others use VAE. Nevertheless, the detectors successfully identified these images, suggesting that different autoencoders introduce similar artifacts into the images.

To evaluate the robustness of the models to image distortions such as JPEG compression or resizing, we conducted additional experiments. The results are shown in Table~\ref{tab:jpeg-results}.

\begin{table}[h]
\centering
\caption{Results with JPEG Compression and Resizing}
\label{tab:jpeg-results}
\resizebox{\columnwidth}{!}{%
\begin{tabular}{@{}llccccc@{}}
\toprule
\multicolumn{1}{c}{\multirow{2}{*}{\textbf{Data Source}}} & \multicolumn{1}{c}{\multirow{2}{*}{\textbf{Model}}} & \multirow{2}{*}{\textbf{Metric}} & \multicolumn{2}{c}{\textbf{JPEG}} & \multicolumn{2}{c}{\textbf{Resize}} \\ \cmidrule(l){4-7} 
\multicolumn{1}{c}{} & \multicolumn{1}{c}{} &  & \textbf{90} & \textbf{80} & \textbf{75\%} & \textbf{50\%} \\ \midrule
\multirow{3}{*}{original} & EfficientNet-V2 B0 & \multirow{3}{*}{FPR} & 0.00141 & 0.00101 & 0.00674 & 0.00824 \\
 & EVA-02 ViT L/14 &  & 0.00098 & 0.00103 & 0.00160 & 0.00204 \\
 & ConvNext Large &  & 0.00022 & 0.00011 & 0.00038 & 0.00043 \\ \midrule
\multirow{3}{*}{high\_res\_imgs} & EfficientNet-V2 B0 & \multirow{3}{*}{FPR} & 0.00000 & 0.00000 & 0.00000 & 0.00000 \\
 & EVA-02 ViT L/14 &  & 0.00000 & 0.00000 & 0.00014 & 0.00000 \\
 & ConvNext Large &  & 0.00000 & 0.00000 & 0.00014 & 0.00000 \\ \midrule
\multirow{3}{*}{\begin{tabular}[c]{@{}l@{}}cagliostrolab\_\\animagine\_xl\_3.1\end{tabular}} & EfficientNet-V2 B0 & \multirow{3}{*}{TPR} & 0.35400 & 0.21600 & 0.72000 & 0.23300 \\
 & EVA-02 ViT L/14 &  & 0.98900 & 0.96800 & 0.69400 & 0.06100 \\
 & ConvNext Large &  & 0.91900 & 0.78500 & 0.41700 & 0.00300 \\ \midrule
\multirow{3}{*}{\begin{tabular}[c]{@{}l@{}}CompVis stable\_\\diffusion\_v1\_4\end{tabular}} & EfficientNet-V2 B0 & \multirow{3}{*}{TPR} & 0.41460 & 0.18950 & 0.31090 & 0.05026 \\
 & EVA-02 ViT L/14 &  & 0.99050 & 0.98210 & 0.6910 & 0.01465 \\
 & ConvNext Large &  & 0.96230 & 0.87320 & 0.25650 & 0.00105 \\ \midrule
\multirow{3}{*}{\begin{tabular}[c]{@{}l@{}}CompVis\_ldm\_\\text2im\_large\_256\end{tabular}} & EfficientNet-V2 B0 & \multirow{3}{*}{TPR} & 0.91500 & 0.91600 & - & - \\
 & EVA-02 ViT L/14 &  & 0.98300 & 0.98300 & - & - \\
 & ConvNext Large &  & 0.99600 & 0.99200 & - & - \\ \midrule
\multirow{3}{*}{\begin{tabular}[c]{@{}l@{}}dreamlike\_art\\\_dreamlike\_\\photoreal\_2.0\end{tabular}} & EfficientNet-V2 B0 & \multirow{3}{*}{TPR} & 0.43400 & 0.22100 & 0.47600 & 0.06700 \\
 & EVA-02 ViT L/14 &  & 1.00000 & 0.99800 & 0.86700 & 0.02300 \\
 & ConvNext Large &  & 0.98400 & 0.86200 & 0.34400 & 0.00100 \\ \midrule
\multirow{3}{*}{\begin{tabular}[c]{@{}l@{}}stabilityai\_\\stable\_diffusion\_2\end{tabular}} & EfficientNet-V2 B0 & \multirow{3}{*}{TPR} & 0.32400 & 0.23800 & 0.36300 & 0.04100 \\
 & EVA-02 ViT L/14 &  & 0.99900 & 0.99700 & 0.94500 & 0.03200 \\
 & ConvNext Large &  & 0.98900 & 0.93300 & 0.61000 & 0.00100 \\ \midrule
\multirow{3}{*}{\begin{tabular}[c]{@{}l@{}}kandinsky\_community\\\_kandinsky\_3\end{tabular}} & EfficientNet-V2 B0 & \multirow{3}{*}{TPR} & 0.40100 & 0.19200 & 0.56200 & 0.25200 \\
 & EVA-02 ViT L/14 &  & 0.98200 & 0.96500 & 0.77800 & 0.15600 \\
 & ConvNext Large &  & 0.9530 & 0.76700 & 0.61800 & 0.01300 \\ \midrule
\multirow{3}{*}{\begin{tabular}[c]{@{}l@{}}stabilityai\_stable\_\\diffusion\_2\_1\end{tabular}} & EfficientNet-V2 B0 & \multirow{3}{*}{TPR} & 0.34400 & 0.23400 & 0.40000 & 0.06500 \\
 & EVA-02 ViT L/14 &  & 1.00000 & 0.99800 & 0.95300 & 0.04700 \\
 & ConvNext Large &  & 0.99200 & 0.93000 & 0.61500 & 0.00200 \\ \midrule
\multirow{3}{*}{\begin{tabular}[c]{@{}l@{}}runwayml\_stable\_\\diffusion\_v1\_5\end{tabular}} & EfficientNet-V2 B0 & \multirow{3}{*}{TPR} & 0.38720 & 0.16590 & 0.31100 & 0.04384 \\
 & EVA-02 ViT L/14 &  & 0.99580 & 0.99260 & 0.67530 & 0.01356 \\
 & ConvNext Large &  & 0.96550 & 0.87890 & 0.25050 & 0.00000 \\ \midrule
\multirow{3}{*}{\begin{tabular}[c]{@{}l@{}}ckpt\_anything\\\_v3.0\end{tabular}} & EfficientNet-V2 B0 & \multirow{3}{*}{TPR} & 0.79800 & 0.68600 & 0.59600 & 0.06800 \\
 & EVA-02 ViT L/14 &  & 1.00000 & 0.99800 & 0.51600 & 0.01900 \\
 & ConvNext Large &  & 0.98600 & 0.95300 & 0.00700 & 0.00100 \\ \midrule
\multirow{3}{*}{\begin{tabular}[c]{@{}l@{}}stabilityai\_stable\_\\diffusion\_xl\_base\_1.0\end{tabular}} & EfficientNet-V2 B0 & \multirow{3}{*}{TPR} & 0.13000 & 0.05200 & 0.44700 & 0.03400 \\
 & EVA-02 ViT L/14 &  & 0.99600 & 0.98300 & 0.73100 & 0.03300 \\
 & ConvNext Large &  & 0.97100 & 0.73200 & 0.26500 & 0.00100 \\ \midrule
\multirow{3}{*}{\begin{tabular}[c]{@{}l@{}}facebook\_DiT\_\\XL\_2\_512\end{tabular}} & EfficientNet-V2 B0 & \multirow{3}{*}{TPR} & 0.41100 & 0.32100 & 0.40100 & 0.23700 \\
 & EVA-02 ViT L/14 &  & 0.89900 & 0.89100 & 0.61500 & 0.11200 \\
 & ConvNext Large &  & 0.90600 & 0.82600 & 0.38400 & 0.06200 \\ \midrule
\multirow{3}{*}{MidJourney\_v6} & EfficientNet-V2 B0 & \multirow{3}{*}{TPR} & 0.47100 & 0.38620 & 0.57170 & 0.07563 \\
 & EVA-02 ViT L/14 &  & 0.97640 & 0.97560 & 0.85470 & 0.02857 \\
 & ConvNext Large &  & 0.94710 & 0.89670 & 0.61710 & 0.00504 \\ \bottomrule
\end{tabular}%
}
\end{table}

We did not apply the resize transformation to the CompVis\_ldm\_text2im\_large\_256 model, as the generated image size is 256x256. Reducing its dimensions further would result in an image smaller than required by our detectors.

The results show that the EVA-02 ViT L/14 model was the most resilient to both types of distortions. ConvNext Large performed well under JPEG compression but showed a significant drop in detection quality when images were resized.

In summary, the ConvNext Large and EVA-02 ViT L/14 models demonstrate strong detection performance for generated images, with the EVA-02 model being particularly robust against image compression and resizing. Meanwhile, EfficientNet-V2 B0 was less effective in detecting both reconstructed and generated images.

\section{Discussion}
\label{sec:disc}
An experiment was conducted to investigate why generated images are relatively easy to detect. We created a simple PNG image with black and white geometric shapes. The image was then reconstructed using autoencoders from latent diffusion models. The image was also compressed using various JPEG compression settings to set a baseline level of artifacts. 
To improve visibility of pixel artifacts, each unique color in the image was replaced with a randomly generated color.

\begin{figure*}[ht]
    \centering
    \subfigure{\includegraphics[width=2.0\columnwidth]{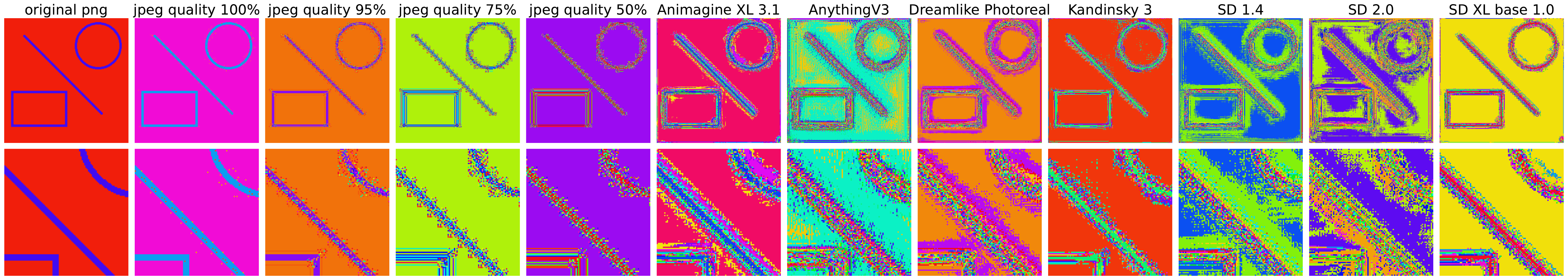}}
    \caption{Visualization results of JPEG (left) and reconstructed (right) images}
    \label{fig:jpeg_reconstr}
\end{figure*}

In Figure \ref{fig:jpeg_reconstr} the visualization results of JPEG and reconstructed images are presented.

The number of unique colors in each image and the percentage of black-and-white pixels are provided in Table \ref{tab:unique_colors}.

\begin{table}[h]
\centering
\caption{Artifacts analysis results}
\label{tab:unique_colors}
\resizebox{\columnwidth}{!}{%
\begin{tabular}{@{}lcccccc@{}}
\toprule
\multicolumn{1}{c}{\multirow{2}{*}{\textbf{Data Source}}} & \multicolumn{3}{c}{\textbf{Unique colors}} & \multicolumn{3}{c}{\textbf{Black-and-white pixels, \%}} \\ \cmidrule(l){2-7} 
\multicolumn{1}{c}{} & \textbf{Default} & \textbf{JPEG 85} & \textbf{Resize 50\%} & \textbf{Default} & \textbf{JPEG 85} & \textbf{Resize 50\%} \\ \midrule
original & 2 & - & - & 1.00000 & - & - \\
JPEG 100 & 4 & - & - & 0.99620 & - & - \\
JPEG 95 & 16 & - & - & 0.95410 & - & - \\
JPEG 75 & 65 & - & - & 0.91820 & - & - \\
JPEG 50 & 121 & - & - & 0.91770 & - & - \\
cagliostrolab\_animagine\_xl\_3.1 & 1424 & 233 & 575 & 0.03857 & 0.05458 & 0.00024 \\
CompVis\_stable\_diffusion\_v1\_4 & 2975 & 895 & 1143 & 0.52060 & 0.76050 & 0.42060 \\
stabilityai\_stable\_diffusion\_2 & 2133 & 436 & 803 & 0.14560 & 0.45920 & 0.06665 \\
stabilityai\_stable\_diffusion\_xl\_base\_1.0 & 1404 & 228 & 569 & 0.05136 & 0.07319 & 0.00116 \\
kandinsky\_community\_kandinsky\_3 & 957 & 187 & 454 & 0.05194 & 0.08134 & 0.00720 \\
ckpt\_anything\_v3.0 & 3645 & 848 & 1113 & 0.00058 & 0.00015 & 0.00000 \\
dreamlike\_art\_dreamlike\_photoreal\_2.0 & 3495 & 1074 & 1124 & 0.64280 & 0.75130 & 0.58500 \\ \bottomrule
\end{tabular}
}
\end{table}

\begin{figure*}[ht]
    \centering
    \subfigure{\includegraphics[width=0.31\textwidth,keepaspectratio]{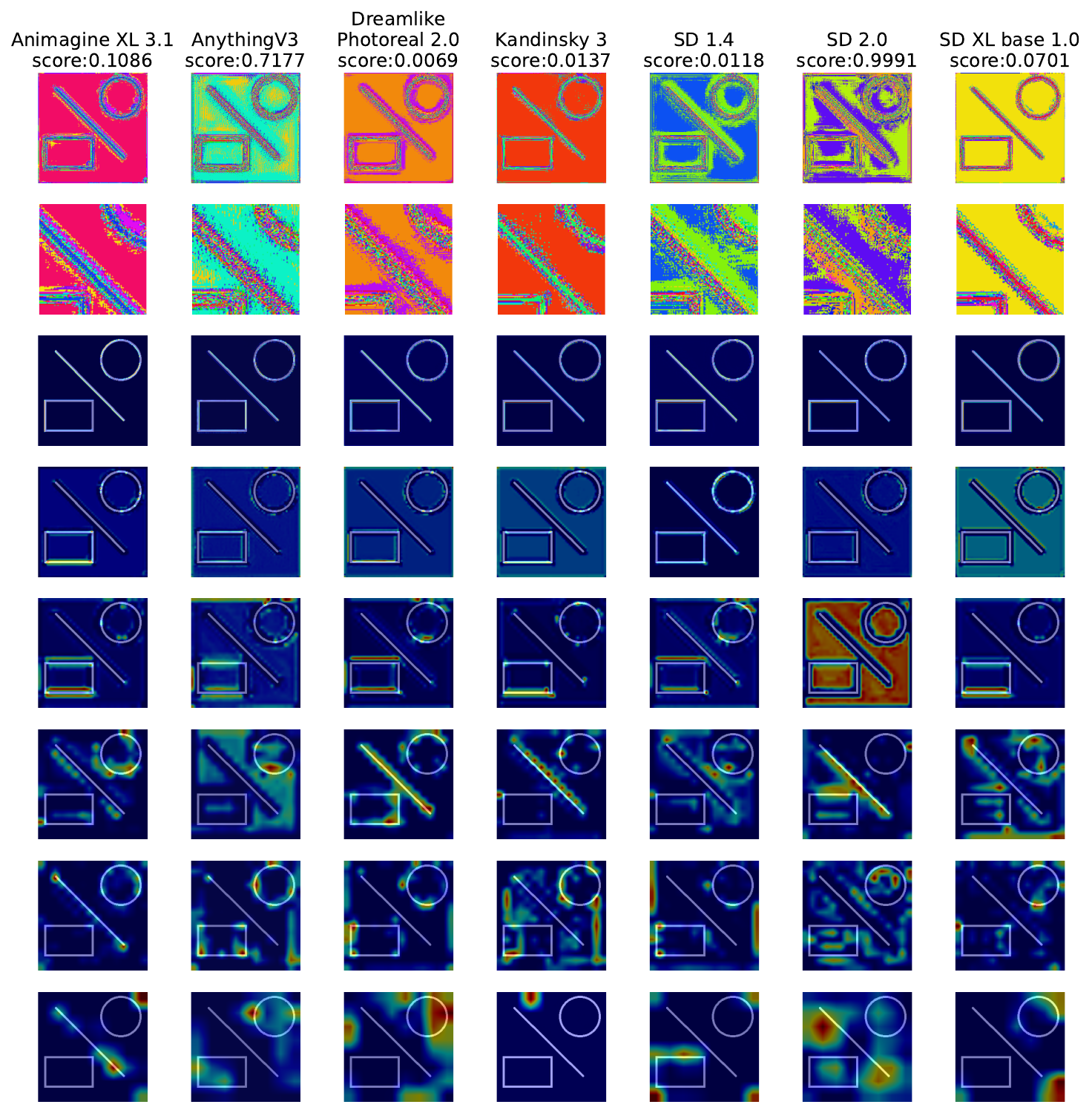}}
    \subfigure{\includegraphics[width=0.34\textwidth, keepaspectratio]{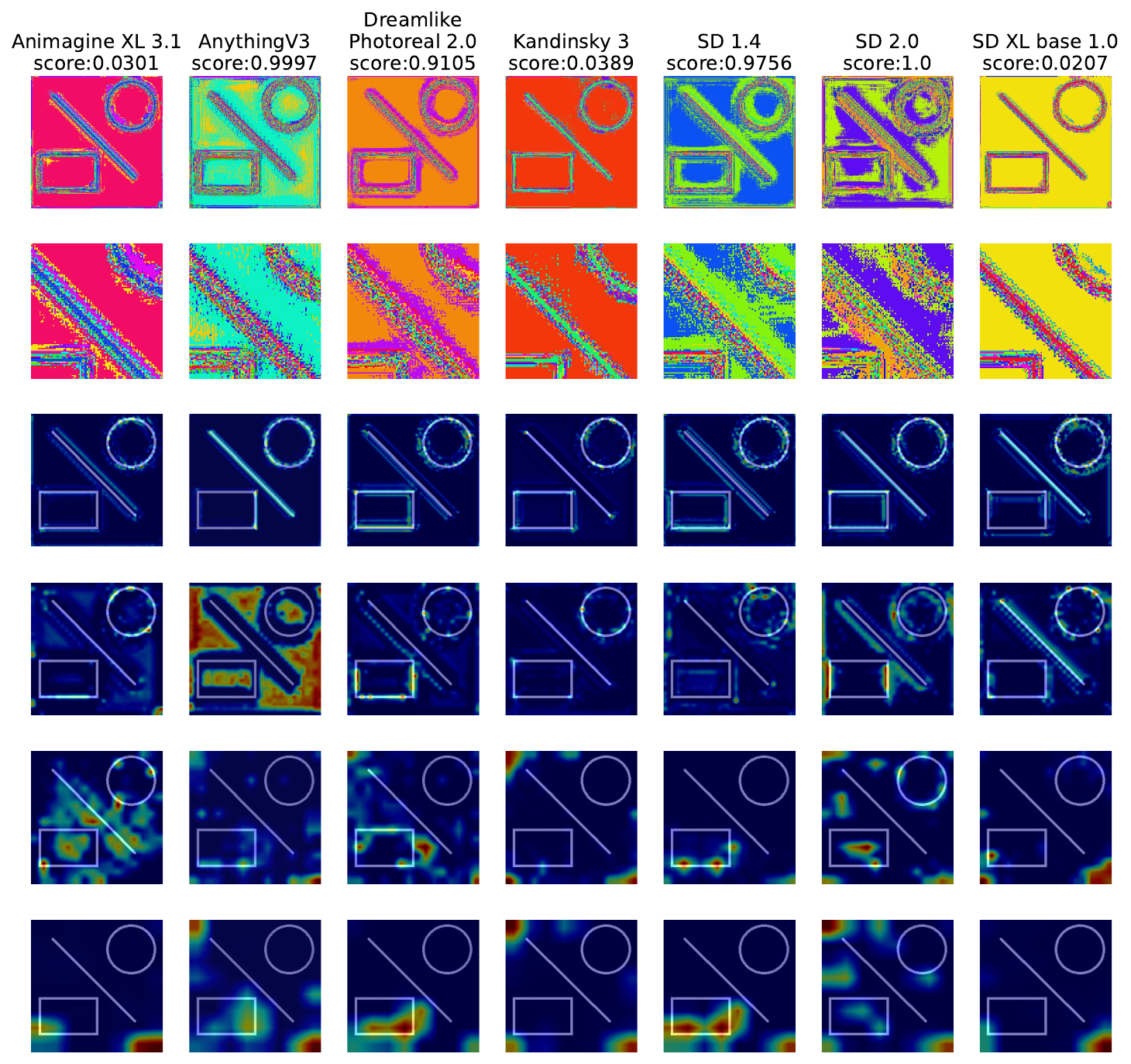}}
    \subfigure{\includegraphics[width=0.33\textwidth, keepaspectratio]{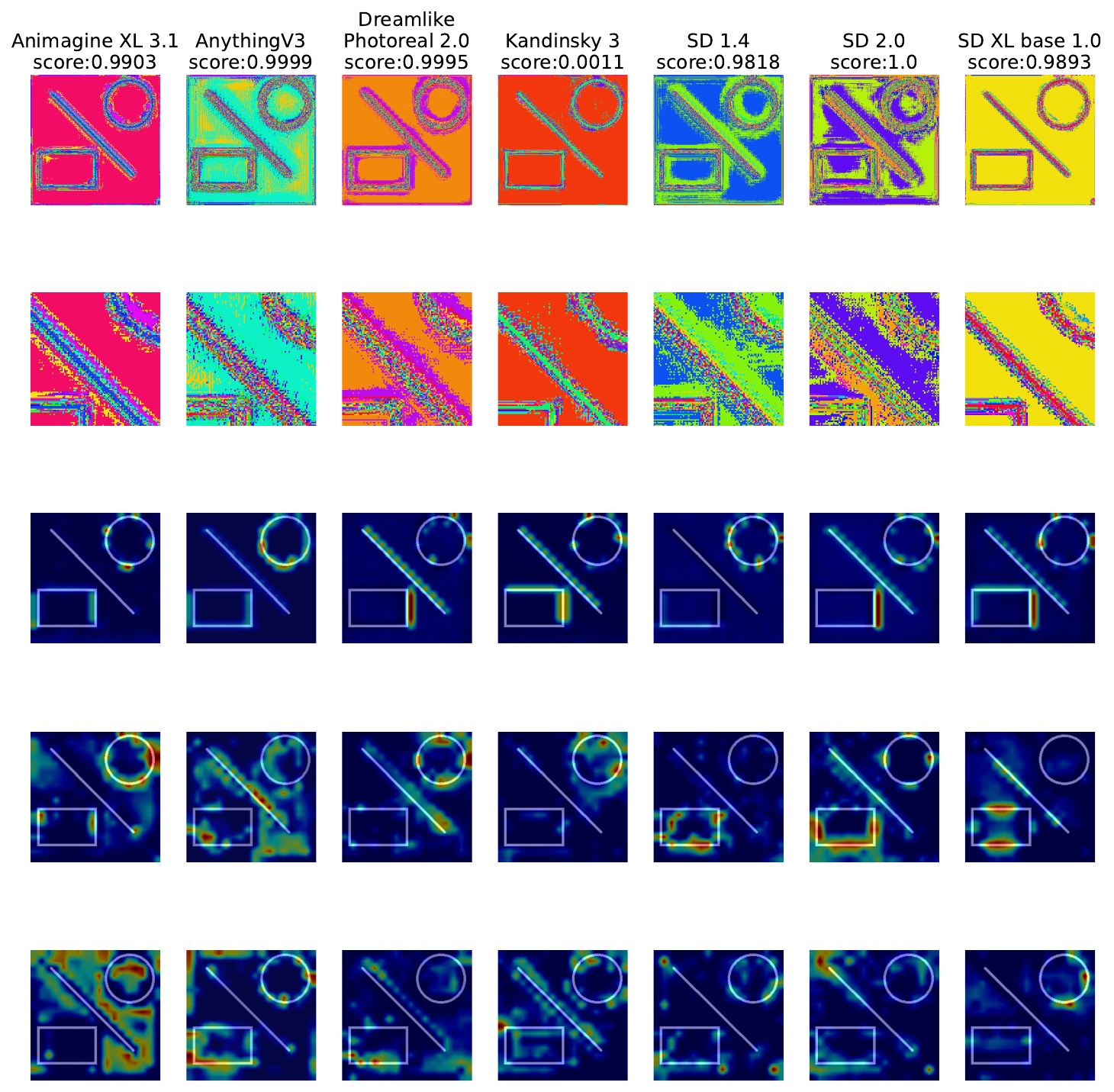}}
    \caption{Visualization results of EfficientNet-V2 B0 (left), ConvNext Large (middle) and EVA-02 ViT L/14 (right)}
    \label{fig:xgradcam}
\end{figure*}

We did not apply any transformations to JPEG images, as they serve as a baseline for the level of introduced artifacts. As shown in the table, autoencoders significantly increase the number of unique colors in an image and greatly reduce the percentage of black-and-white pixels, indicating that they introduce substantial distortions to the original images that can be detected by classifiers. Using XGradCAM\cite{fu2020axiom} visualization, we examined which image regions detectors prioritize when identifying reconstructions produced by autoencoders. This tool highlights the areas most critical for the detector’s classification decisions, see Figure \ref{fig:xgradcam}.

For ConvNext Large, the final layer of each stage was visualized; for EfficientNet-V2 B0, the final layer within each block; and for EVA-02 ViT L/14, the normalization layers in the first, middle, and final blocks.

From the figure, we can conclude that the models focus on anomalous areas where there is a concentration of pixels of various colors. We believe that it is caused by the noise introduced by autoencoders.

Next, we compared our method with the state-of-the-art approaches. For this comparison, we used a dataset combining the original, high\_res\_imgs, and generated images datasets.
The metrics used for comparison are TPR @0.1\% FPR and AUC ROC. The testing results are presented in Table \ref{tab:testing_w_others}.

\begin{table}[h]
\centering
\caption{Comparison with others (higher is better)}
\label{tab:testing_w_others}
\begin{tabular}{@{}ccc@{}}
\toprule
\textbf{Model} & \textbf{TPR @0.1\% FPR} & \textbf{AUC ROC} \\ \midrule
\cite{ricker2024aeroblade} & 0.0000 & 0.5373 \\
\cite{corvi2023detection} & 0.8573 & 0.9895 \\
\cite{ojha2023towards} & 0.0131 & 0.6041 \\
EfficientNet-V2 B0 & 0.4063 & 0.9590 \\
ConvNext Large & 0.9528 & 0.9974 \\
EVA-02 ViT L/14 & 0.9236 & 0.9962 \\ \bottomrule
\end{tabular}
\end{table}

The testing results demonstrate that our method achieves a high detection rate of generated images while minimizing the rate of false positives, making it suitable for real-world applications. The results also confirm that it is not necessary to train on generated images to detect them effectively.

\section{Conclusion}
\label{sec:concl}
In this paper, we proposed an efficient method for detecting images generated using latent diffusion models (LDM). The main advantage of the proposed approach is that it does not require training on generated images, but instead uses the artifacts introduced by the autoencoders of these models. Experimental results demonstrate the method's high generalization ability and robustness to various distortions such as JPEG compression and resizing.

The key takeaway from this work is that the artifacts introduced by LDM autoencoders are similar regardless of the specific model used, allowing for effective detection of images created by various diffusion models. The method showed strong results across different architectures, including Stable Diffusion and DiT.

In future work, we plan to enhance the method’s robustness to new types of distortions, as well as adapt it to other diffusion model architectures. Additionally, we aim to focus on developing methods that provide human-readable explanations, enabling better interpretation of the features used to detect generated or reconstructed images.

\section*{Acknowledgments}
The reported study was partially funded by the budget project FFZF-2022-0007.

\bibliographystyle{IEEEtran}
\bibliography{full}

\begin{thebibliography}{10}
\providecommand{\url}[1]{#1}
\csname url@samestyle\endcsname
\providecommand{\newblock}{\relax}
\providecommand{\bibinfo}[2]{#2}
\providecommand{\BIBentrySTDinterwordspacing}{\spaceskip=0pt\relax}
\providecommand{\BIBentryALTinterwordstretchfactor}{4}
\providecommand{\BIBentryALTinterwordspacing}{\spaceskip=\fontdimen2\font plus
\BIBentryALTinterwordstretchfactor\fontdimen3\font minus \fontdimen4\font\relax}
\providecommand{\BIBforeignlanguage}[2]{{%
\expandafter\ifx\csname l@#1\endcsname\relax
\typeout{** WARNING: IEEEtran.bst: No hyphenation pattern has been}%
\typeout{** loaded for the language `#1'. Using the pattern for}%
\typeout{** the default language instead.}%
\else
\language=\csname l@#1\endcsname
\fi
#2}}
\providecommand{\BIBdecl}{\relax}
\BIBdecl

\bibitem{ho2020denoising}
J.~Ho, A.~Jain, and P.~Abbeel, ``Denoising diffusion probabilistic models,'' \emph{Advances in neural information processing systems}, vol.~33, pp. 6840--6851, 2020.

\bibitem{songdenoising}
J.~Song, C.~Meng, and S.~Ermon, ``Denoising diffusion implicit models,'' in \emph{International Conference on Learning Representations}.

\bibitem{dhariwal2021diffusion}
P.~Dhariwal and A.~Nichol, ``Diffusion models beat gans on image synthesis,'' \emph{Advances in neural information processing systems}, vol.~34, pp. 8780--8794, 2021.

\bibitem{rombach2022high}
R.~Rombach, A.~Blattmann, D.~Lorenz, P.~Esser, and B.~Ommer, ``High-resolution image synthesis with latent diffusion models,'' in \emph{Proceedings of the IEEE/CVF conference on computer vision and pattern recognition}, 2022, pp. 10\,684--10\,695.

\bibitem{schuhmann2022laion}
C.~Schuhmann, R.~Beaumont, R.~Vencu, C.~Gordon, R.~Wightman, M.~Cherti, T.~Coombes, A.~Katta, C.~Mullis, M.~Wortsman \emph{et~al.}, ``Laion-5b: An open large-scale dataset for training next generation image-text models,'' \emph{Advances in Neural Information Processing Systems}, vol.~35, pp. 25\,278--25\,294, 2022.

\bibitem{comfyanonymous2024comfyui}
\BIBentryALTinterwordspacing
ComfyAnonymous, ``Comfyui,'' 2024, accessed: 2024-10-31. [Online]. Available: \url{https://github.com/comfyanonymous/ComfyUI}
\BIBentrySTDinterwordspacing

\bibitem{automatic11112022stable}
\BIBentryALTinterwordspacing
AUTOMATIC1111, ``Stable diffusion webui,'' 2024, accessed: 2024-10-31. [Online]. Available: \url{https://github.com/AUTOMATIC1111/stable-diffusion-webui}
\BIBentrySTDinterwordspacing

\bibitem{openai_dalle_2022}
\BIBentryALTinterwordspacing
OpenAI, ``{DALL·E now available without waitlist},'' 2022, accessed: 2024-10-20. [Online]. Available: \url{https://openai.com/index/dall-e-now-available-without-waitlist/}
\BIBentrySTDinterwordspacing

\bibitem{adobe_firefly_2023}
\BIBentryALTinterwordspacing
Adobe, ``The future is firefly: Unlock new levels of creativity with the latest generative ai innovations,'' 2023, accessed: 2024-10-20. [Online]. Available: \url{https://blog.adobe.com/en/publish/2023/10/10/future-is-firefly-adobe-max}
\BIBentrySTDinterwordspacing

\bibitem{ronneberger2015u}
O.~Ronneberger, P.~Fischer, and T.~Brox, ``U-net: Convolutional networks for biomedical image segmentation,'' in \emph{Medical image computing and computer-assisted intervention--MICCAI 2015: 18th international conference, Munich, Germany, October 5-9, 2015, proceedings, part III 18}.\hskip 1em plus 0.5em minus 0.4em\relax Springer, 2015, pp. 234--241.

\bibitem{peebles2023scalable}
W.~Peebles and S.~Xie, ``Scalable diffusion models with transformers,'' in \emph{Proceedings of the IEEE/CVF International Conference on Computer Vision}, 2023, pp. 4195--4205.

\bibitem{ha2023robust}
H.~Ha, M.~Kim, S.~Han, and S.~Lee, ``Robust deepfake detection method based on ensemble of vit and cnn,'' in \emph{Proceedings of the 38th ACM/SIGAPP Symposium on Applied Computing}, 2023, pp. 1092--1095.

\bibitem{aghasanli2023interpretable}
A.~Aghasanli, D.~Kangin, and P.~Angelov, ``Interpretable-through-prototypes deepfake detection for diffusion models,'' in \emph{Proceedings of the IEEE/CVF international conference on computer vision}, 2023, pp. 467--474.

\bibitem{corvi2023detection}
R.~Corvi, D.~Cozzolino, G.~Zingarini, G.~Poggi, K.~Nagano, and L.~Verdoliva, ``On the detection of synthetic images generated by diffusion models,'' in \emph{ICASSP 2023-2023 IEEE International Conference on Acoustics, Speech and Signal Processing (ICASSP)}.\hskip 1em plus 0.5em minus 0.4em\relax IEEE, 2023, pp. 1--5.

\bibitem{ojha2023towards}
U.~Ojha, Y.~Li, and Y.~J. Lee, ``Towards universal fake image detectors that generalize across generative models,'' in \emph{Proceedings of the IEEE/CVF Conference on Computer Vision and Pattern Recognition}, 2023, pp. 24\,480--24\,489.

\bibitem{radford2021learning}
A.~Radford, J.~W. Kim, C.~Hallacy, A.~Ramesh, G.~Goh, S.~Agarwal, G.~Sastry, A.~Askell, P.~Mishkin, J.~Clark \emph{et~al.}, ``Learning transferable visual models from natural language supervision,'' in \emph{International conference on machine learning}.\hskip 1em plus 0.5em minus 0.4em\relax PMLR, 2021, pp. 8748--8763.

\bibitem{ricker2022towards}
J.~Ricker, S.~Damm, T.~Holz, and A.~Fischer, ``Towards the detection of diffusion model deepfakes,'' \emph{arXiv preprint arXiv:2210.14571}, 2022.

\bibitem{wang2023dire}
Z.~Wang, J.~Bao, W.~Zhou, W.~Wang, H.~Hu, H.~Chen, and H.~Li, ``Dire for diffusion-generated image detection,'' in \emph{Proceedings of the IEEE/CVF International Conference on Computer Vision}, 2023, pp. 22\,445--22\,455.

\bibitem{ricker2024aeroblade}
J.~Ricker, D.~Lukovnikov, and A.~Fischer, ``Aeroblade: Training-free detection of latent diffusion images using autoencoder reconstruction error,'' in \emph{Proceedings of the IEEE/CVF Conference on Computer Vision and Pattern Recognition}, 2024, pp. 9130--9140.

\bibitem{zhang2018unreasonable}
R.~Zhang, P.~Isola, A.~A. Efros, E.~Shechtman, and O.~Wang, ``The unreasonable effectiveness of deep features as a perceptual metric,'' in \emph{Proceedings of the IEEE conference on computer vision and pattern recognition}, 2018, pp. 586--595.

\bibitem{Gustavosta_Stable_Diffusion_Prompts}
G.~Santana, ``Stable diffusion prompts dataset,'' \url{https://huggingface.co/datasets/Gustavosta/Stable-Diffusion-Prompts}, 2022, accessed: 2024-10-20.

\bibitem{podell2023sdxl}
D.~Podell, Z.~English, K.~Lacey, A.~Blattmann, T.~Dockhorn, J.~M{\"u}ller, J.~Penna, and R.~Rombach, ``Sdxl: Improving latent diffusion models for high-resolution image synthesis,'' \emph{arXiv preprint arXiv:2307.01952}, 2023.

\bibitem{liu2022convnet}
Z.~Liu, H.~Mao, C.-Y. Wu, C.~Feichtenhofer, T.~Darrell, and S.~Xie, ``A convnet for the 2020s,'' in \emph{Proceedings of the IEEE/CVF conference on computer vision and pattern recognition}, 2022, pp. 11\,976--11\,986.

\bibitem{fang2024eva}
Y.~Fang, Q.~Sun, X.~Wang, T.~Huang, X.~Wang, and Y.~Cao, ``Eva-02: A visual representation for neon genesis,'' \emph{Image and Vision Computing}, vol. 149, p. 105171, 2024.

\bibitem{tan2021efficientnetv2}
M.~Tan and Q.~Le, ``Efficientnetv2: Smaller models and faster training,'' in \emph{International conference on machine learning}.\hskip 1em plus 0.5em minus 0.4em\relax PMLR, 2021, pp. 10\,096--10\,106.

\bibitem{zheng2022movq}
C.~Zheng, T.-L. Vuong, J.~Cai, and D.~Phung, ``Movq: Modulating quantized vectors for high-fidelity image generation,'' \emph{Advances in Neural Information Processing Systems}, vol.~35, pp. 23\,412--23\,425, 2022.

\bibitem{fu2020axiom}
R.~Fu, Q.~Hu, X.~Dong, Y.~Guo, Y.~Gao, and B.~Li, ``Axiom-based grad-cam: Towards accurate visualization and explanation of cnns,'' \emph{arXiv preprint arXiv:2008.02312}, 2020.

\end{thebibliography}

\vfill

\end{document}